\documentclass[conference]{IEEEtran}
\usepackage{times}

\usepackage{multicol}
\usepackage[bookmarks=true]{hyperref}

\usepackage{graphics} 
\usepackage{graphicx}
\usepackage{todonotes}

\usepackage{cite}

\title{\LARGE \bf
Practical Robot Learning from Demonstrations using\\ Deep End-to-End Training
}

\author{Akansel Cosgun$^{1}$,\thanks{Authors are affliated with ARC Centre for Excellence for Robot Vision} \thanks{$^{1}$Authors are with the Monash University, Australia}Thomas Rowntree$^{2}$, Ian Reid$^{2}$ and Tom Drummond$^{1}$ \thanks{$^{2}$Authors are with the University of Adelaide, Australia}}

\IEEEoverridecommandlockouts

\begin{document}

\maketitle

\begin{abstract}
 Robots need to learn behaviors in intuitive and practical ways for widespread deployment in human environments. To learn a robot behavior end-to-end, we train a variant of the ResNet that maps eye-in-hand camera images to end-effector velocities. In our setup, a human teacher demonstrates the task via joystick. We show that a simple servoing task can be learned in less than an hour including data collection, model training and deployment time. Moreover, 16 minutes of demonstrations were enough for the robot to learn the task.
\end{abstract}

\IEEEpeerreviewmaketitle

\section{Introduction}
\label{sec:introduction}


Robots that operate in human environments have potential for making our lives easier. Unlike industrial environments, fixed robot trajectories are not usable in human environments since the objects and situations that the robot can encounter are virtually unbounded. Robots should be able to adapt to ever-changing situations and learn new skills to operate in unstructured environments. Utilizing recent machine learning techniques is a promising avenue for robotics.

Robotics researchers have increasingly been using deep learning, which emerged as a powerful statistical tool for processing high-dimensional inputs \cite{krizhevsky2012imagenet}. Data for robot learning can be gathered in a multitude of ways including trial-and-error, from human experts and by observations \cite{hussein2017imitation}. Reinforcement Learning (RL) is the most popular trial-and-error method where the robot explores the state space itself \cite{zhang2015towards}. Using reinforcement learning on real robots is impractical because the robot would need thousands of trials and it is challenging to explore the state space in a safe manner. Transferring RL policies from simulation to real robot is a solution proposed for this problem \cite{andrychowicz2018learning}. Human expert methods include Learning from Demonstration (LfD) in which the robot learns the task by observing how humans behave in the same task \cite{akgun2012keyframe}. Observation methods \cite{murali2015learning} often involve learning by watching either an expert or another agent execute the task. How much human demonstration data a learning algorithm needs is an important criteria.

In this work, we use end-to-end learning from human demonstration data for data efficient behavior learning. The contribution of this work is a practical pipeline which takes a short amount of demonstration time to teach new behaviors to a robot. This approach provides a data-efficient alternative to reinforcement learning for robots. It is valuable to teach skills to robot in a short amount of time so that the outcome policies can be tested quickly and more data can be collected if necessary.


\begin{figure}[t!]
\centering
\includegraphics[trim=0.5cm 1cm 5cm 0, clip, width=0.41\textwidth]{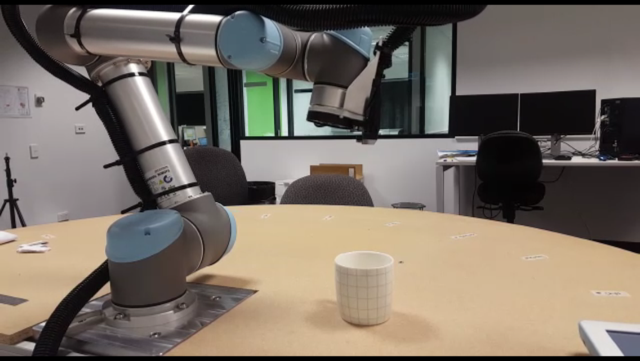}
\caption{The robot was able to learn visual servoing to a cup after only 16 minutes of demonstration time from a human teacher.}
\label{fig:intro}
\end{figure}

\section{End-to-End Training}
\label{sec:approach}

We use an eye-in-hand setup where a RGB camera is attached to the end-effector. Our approach is based on mapping the input image to the end-effector velocity. A Convolutional Neural Network (CNN) is used to represent mapping function. End-to-end training shows promise in learning visuomotor policies~\cite{levine2016end}.

The data collection is achieved by the human providing demonstrations of the intended behavior using a joystick. The operator uses start/stop recording buttons to initiate and pause recording of demonstration data. The human teacher aims for demonstrations with different initial states since it helps explore the state space. The recorded data consists of the raw image and the end-effector velocity (in the end-effector coordinate frame) that the operator applied.

The CNN represents the mapping from the input image to the end-effector velocity. The input images are re-sized to 224 by 224 pixels. We use a modified version of the ResNet architecture~\cite{he2016deep} where the global average pooling and fully connected layers are removed and replaced by a new 1x1 convolutional layer for channel reduction, and new fully connected layers. Instead of outputting class probabilities as the original ResNet, we output 6 floating point numbers which represent the twist of the end-effector in the end-effector coordinate frame: $t_{network}=(v_x,v_y,v_z,\omega_x,\omega_y,\omega_z)$.
We use pretrained weights for the original ResNet layers and randomly initialised weights for the new layers. All weights are fine-tuned during training. These modifications makes the neural net training much faster (less than 30 minutes for the servoing task), which makes it feasible to learn robot behaviors practically.

During robot execution, the camera image is resized to 224 by 224 pixels and fed to the network for a feed-forward pass-through. To ensure that the robot does not mistakenly collide with itself or the table, we compute a safety twist $t_{safety}$ at every frame. We apply the aggregated twist $t_{final} = t_{safety} + t_{network}$ to the robot using the velocity control feature of the UR robot drivers. To compute $t_{safety}$, we construct a virtual workspace volume from spheres, cylinders and planes. If the end-effector position is inside this work area, then $t_{safety}=0$. Otherwise, we compute the twist term that will direct the end-effector back inside the work area with a magnitude that is proportional to the distance to the work area boundary. We were able to execute the approach in real-time, at the rate camera images are received (30 Hz).

\section{Experiments}
\label{sec:evaluation}

The evaluation of our approach is focused on how much human demonstration effort is required for the robot to learn a new task. We chose a visual servoing task in order to validate the soundness and practicality of the approach. A human teacher (one of the authors) demonstrated the task for 20 minutes with different initial object and robot arm positions, ensuring that the robot would point directly above the cup with the camera looking straight into it. We measure the demonstration time in  wall-clock time: we think it represents a realistic estimate of how much time it would take a robot teacher to demonstrate a task. We consider five checkpoints in this dataset: 4,8,12,16 and 20 minute marks as seen in Table~\ref{tab:cup-training-set}. For example, for the 4 minutes dataset we use the first four minutes of demonstrations only. Note that the frame count is not an exact multiple of demonstration time. This is because some of the demonstration time is not recorded while manually moving the robot and or the cup.

\begin{table}[t!]
\centering
\begin{tabular}{|c|c|c|}
\hline
Training Set & \multicolumn{1}{c|}{\begin{tabular}[c]{@{}c@{}}Demo Time\\ (Minutes)\end{tabular}} & \multicolumn{1}{c|}{Num. Images} \\ \hline
cup-4 & 4 & 672 \\ \hline
cup-8 & 8 & 1306 \\ \hline
cup-12 & 12 & 1997 \\ \hline
cup-16 & 16 & 2686 \\ \hline
cup-20 & 20 & 3399 \\ \hline
\end{tabular}
\caption{The split datasets used in experiments}
\vspace{-0.5cm}
\label{tab:cup-training-set}
\end{table}

\begin{figure}[h!]
\centering
\includegraphics[trim=0.5cm 0.5cm 1cm 0.5cm, clip, width=0.4\textwidth]{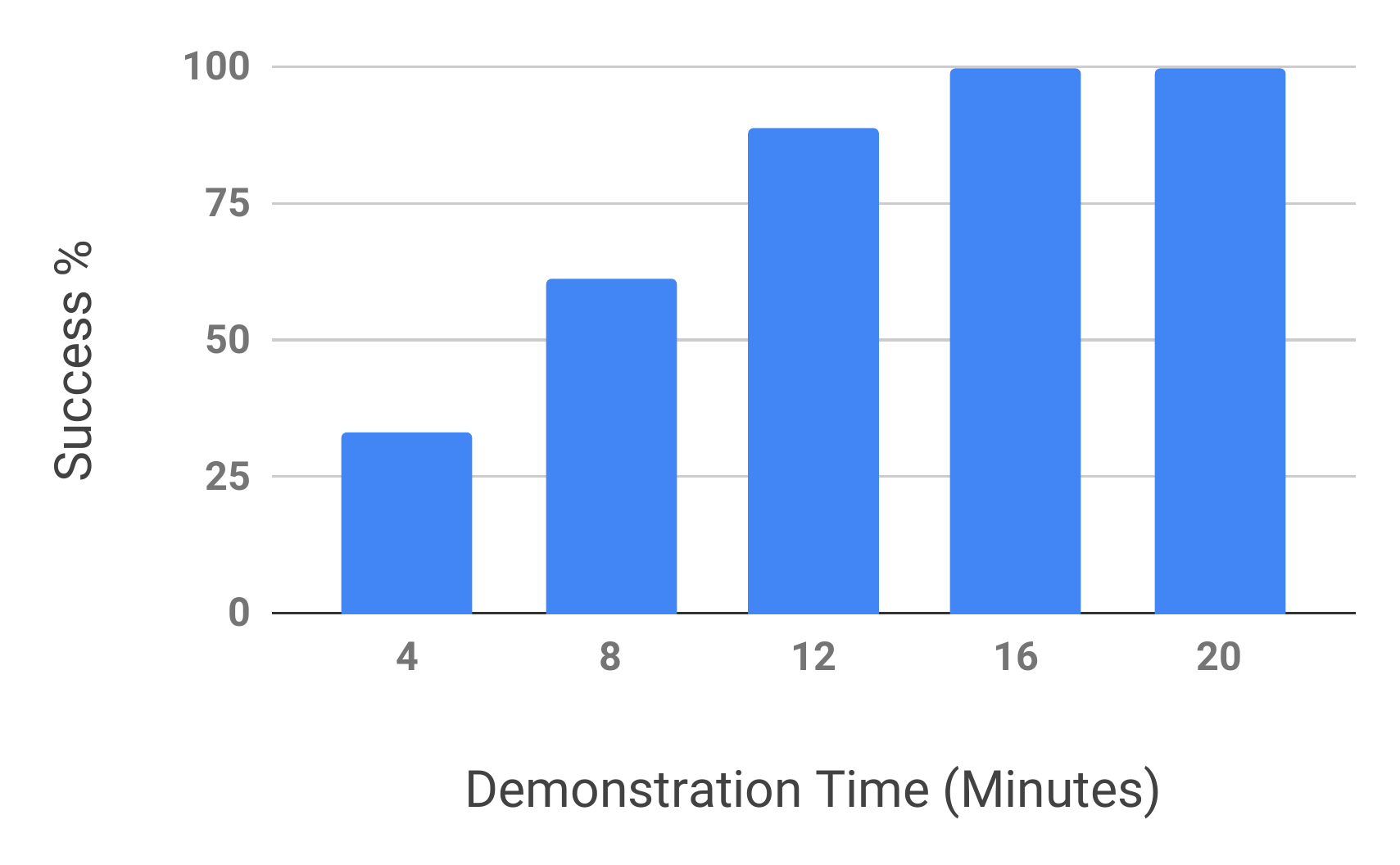}
\caption{Success rate of the servoing task with varying amounts of human demonstrations}
\label{fig:cup-success}
\end{figure}

The success metric for the experiments is whether the robot can get near to the goal position (above the cup) under 1 minute. For each dataset, we kept the cup position constant and varied the initial robot position. We used 9 initial positions where 5 of them were close to the cup and 4 were relatively further. For each initial position, we run two trials. A total of 90 trials were executed where we recorded the success rate. The results are shown in Figure~\ref{fig:cup-success}.

The success rate was the lowest for 4-minute dataset and steadily increased as demonstration time went up, as expected. 16 and 20 minute datasets achieved 100\% success. From this result we can claim that 16 minutes of human demonstrations were enough to learn this task. We did not observe catastrophic forgetting for the 20-minute dataset, which is common in neural nets for multi-task learning~\cite{isele2018selective}.

\section{Conclusion and Future Work}
\label{sec:conclusion}

This study shows that the robot can learn a simple task in under 1 hour including demonstration, model training and deployment time. Our approach can also be used for quickly learning an initial policy for data-inefficient reinforcement learning approaches~\cite{brys2015reinforcement}.

Future work includes extending the approach to contact-rich and multi-step tasks. This would require utilizing multi-modal sensory input including images from different cameras and force/torque data.

\bibliographystyle{is-unsrt}
\bibliography{main}

\begin{thebibliography}{10}
\ifx \showCODEN  \undefined \def \showCODEN #1{CODEN #1}  \fi
\ifx \showISBN   \undefined \def \showISBN  #1{ISBN #1}   \fi
\ifx \showISSN   \undefined \def \showISSN  #1{ISSN #1}   \fi
\ifx \showLCCN   \undefined \def \showLCCN  #1{LCCN #1}   \fi
\ifx \showPRICE  \undefined \def \showPRICE #1{#1}        \fi
\ifx \showURL    \undefined \def \showURL {URL }          \fi
\ifx \path       \undefined \input path.sty               \fi
\ifx \ifshowURL \undefined
     \newif \ifshowURL
     \showURLtrue
\fi

\bibitem{krizhevsky2012imagenet}
Alex Krizhevsky, Ilya Sutskever, and Geoffrey~E Hinton.
\newblock Imagenet classification with deep convolutional neural networks.
\newblock In {\em Advances in Neural Information Processing Systems (NIPS)},
  2012.

\bibitem{hussein2017imitation}
Ahmed Hussein, Mohamed~Medhat Gaber, Eyad Elyan, and Chrisina Jayne.
\newblock Imitation learning: A survey of learning methods.
\newblock {\em ACM Computing Surveys (CSUR)}, 50\penalty0 (2):\penalty0 21,
  2017.

\bibitem{zhang2015towards}
Fangyi Zhang, J{\"u}rgen Leitner, Michael Milford, Ben Upcroft, and Peter
  Corke.
\newblock Towards vision-based deep reinforcement learning for robotic motion
  control.
\newblock {\em arXiv preprint arXiv:1511.03791}, 2015.

\bibitem{andrychowicz2018learning}
Marcin Andrychowicz, Bowen Baker, Maciek Chociej, Rafal Jozefowicz, Bob McGrew,
  Jakub Pachocki, Arthur Petron, Matthias Plappert, Glenn Powell, Alex Ray,
  et~al.
\newblock Learning dexterous in-hand manipulation.
\newblock {\em arXiv preprint arXiv:1808.00177}, 2018.

\bibitem{akgun2012keyframe}
Baris Akgun, Maya Cakmak, Karl Jiang, and Andrea~L Thomaz.
\newblock Keyframe-based learning from demonstration.
\newblock {\em International Journal of Social Robotics}, 2012.

\bibitem{murali2015learning}
Adithyavairavan Murali, Siddarth Sen, Ben Kehoe, Animesh Garg, Seth McFarland,
  Sachin Patil, W~Douglas Boyd, Susan Lim, Pieter Abbeel, and Ken Goldberg.
\newblock Learning by observation for surgical subtasks: Multilateral cutting
  of 3d viscoelastic and 2d orthotropic tissue phantoms.
\newblock In {\em 2015 IEEE International Conference on Robotics and Automation
  (ICRA)}.

\bibitem{levine2016end}
Sergey Levine, Chelsea Finn, Trevor Darrell, and Pieter Abbeel.
\newblock End-to-end training of deep visuomotor policies.
\newblock {\em The Journal of Machine Learning Research}, 17\penalty0
  (1):\penalty0 1334--1373, 2016.

\bibitem{he2016deep}
Kaiming He, Xiangyu Zhang, Shaoqing Ren, and Jian Sun.
\newblock Deep residual learning for image recognition.
\newblock In {\em Proceedings of the IEEE conference on computer vision and
  pattern recognition}, 2016.

\bibitem{isele2018selective}
David Isele and Akansel Cosgun.
\newblock Selective experience replay for lifelong learning.
\newblock In {\em Thirty-Second AAAI Conference on Artificial Intelligence},
  2018.

\bibitem{brys2015reinforcement}
Tim Brys, Anna Harutyunyan, Halit~Bener Suay, Sonia Chernova, Matthew~E Taylor,
  and Ann Now{\'e}.
\newblock Reinforcement learning from demonstration through shaping.
\newblock In {\em Twenty-Fourth International Joint Conference on Artificial
  Intelligence}, 2015.

\end{thebibliography}

\end{document}